\documentclass{article}
\usepackage[a4paper,
            bindingoffset=0.2in,
            left=1in,
            right=1in,
            top=1in,
            bottom=1in,
            footskip=.25in]{geometry}
\usepackage{multirow}

\usepackage{cite}
\usepackage[pdftex]{graphicx}
\usepackage{amsmath,amssymb}
\usepackage{hhline}
\usepackage{MnSymbol}
\usepackage[most]{tcolorbox}

\newcommand\erf{\mathrm{erf}}

\DeclareMathOperator*{\argmin}{arg\,min}

\usepackage{authblk}

\title{Exploration, Path Planning with Obstacle and Collision Avoidance in a Dynamic Environment}
\author[1]{Saeid Alirezazadeh}
\author[2]{Lu\'{i}s A. Alexandre}
\affil[1]{C4 - Cloud Computing Competence Centre (C4-UBI), Universidade da Beira Interior, Rua Marqu\^{e}s d'\'{A}vila e Bolama, 6201-001, Covilh\~{a}, Portugal}
\affil[2]{NOVA LINCS, Universidade da Beira Interior, Covilh\~{a}, Portugal}
\affil[1]{Email: saeid.alirezazadeh@gmail.com}

\date{}

\providecommand{\keywords}[1]{\textbf{\textit{Keywords---}} #1}
\begin{document}

\maketitle

\begin{abstract}
If we give a robot the task of moving an object from its current position to another location in an unknown environment, the robot must explore the map, identify all types of obstacles, and then determine the best route to complete the task. We proposed a mathematical model to find an optimal path planning that avoids collisions with all static and moving obstacles and has the minimum completion time and the minimum distance traveled. In this model, the bounding box around obstacles and robots is not considered, so the robot can move very close to the obstacles without colliding with them. We considered two types of obstacles: deterministic, which include all static obstacles such as walls that do not move and all moving obstacles whose movements have a fixed pattern, and non-deterministic, which include all obstacles whose movements can occur in any direction with some probability distribution at any time. We also consider the acceleration and deceleration of the robot to improve collision avoidance.
\end{abstract}

\keywords{Path planning, Collision avoidance, Dynamic environment, Wheeled robot.}

%

\section{Introduction}
Robotic systems are used in people's daily lives. They occur in various contexts, such as industry and manufacturing \cite{ifr:2020, Ross:2017}, military \cite{nath:2014, springer:2013}, household \cite{Prassler:2008, xu:2014}, and others \cite{bruno:2016}. Robotic systems can be classified as single robot or multi-robot. In this paper, we mainly focus on finding the optimal motion of a single robot in a dynamic environment.

Some of the problems of moving robots are path planning, collision avoidance, and map exploration, which are studied independently in the literature. In particular, when a robot is introduced to a new environment and given the task of moving an object from its position to another, the robot should explore the environment while moving to its final position and simultaneously detect and avoid all obstacles, including static and moving obstacles. To avoid obstacles, the robot should find a new route, but changing the robot's speed has been shown to be a better way to avoid collisions, \cite{Foka:2003}. Also, the movement of obstacles should be predicted, and the robot should adjust its movement according to the movement of obstacles nearby. In addition, the cost of the robot to complete the task should be minimized, i.e., the robot should complete the task in the shortest time and travel the shortest distance. Some studies consider a bounding box around an obstacle that a robot should not cross to avoid collisions, \cite{Xin:2018}. However, using bounding boxes is equivalent to reshaping obstacles and requires some estimation, and crossing a bounding box does not necessarily mean that the robot will collide with the obstacle, so the distance traveled and time may not be minimized. This is a complete list of the difficulties in studying robot's motion planning. As far as we know, there is no study that solves all these components simultaneously.

We have developed a mathematical model that can be used for exploration, path planning, and collision avoidance while predicting the obstacle's motion, controlling the robot's velocity, and ignoring bounding boxes. We show through designs that the proposed model minimizes the total distance traveled and the time to complete the task.

The paper is organized as follows. Section 2 discusses related work on motion planning and collision avoidance. Section 3 describes the mathematical model for optimal robot motion planning. And finally, Section 4 presents some conclusions and future lines of work.

\section{Related Works}

\begin{figure}[!h]\centering
\includegraphics[width=0.45\linewidth]{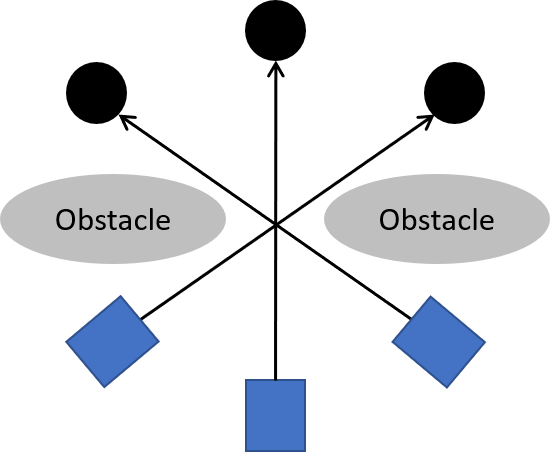}
\caption{Example of multiple vehicles arriving at different positions.}
\label{fig1}
\end{figure}

\cite{Hausler:2016} studied the motion planning of multiple vehicles in a simultaneous arrival problem with collision avoidance and minimization of total energy consumption by minimizing the temporal and spatial constraints. They consider the motions as a linear path between initial and final positions as straight lines. The total motion of each vehicle is converted into segments with elementary linear motions with constant velocities in each segment. For collision avoidance, they consider virtual circles around vehicles and obstacles. A vehicle may collide with obstacles or other vehicles if their virtual circles intersect. Then they translate their model into a multi-objective optimization where they find the maximum velocity vectors of the vehicles to find the optimal solution. In their formulation, the energy consumption is considered as the movement of the vehicles. Their model is a centralized optimization model and does not consider the communication between the vehicles. In their model, it is assumed that all vehicles arrive at the same final position. Figure \ref{fig1} shows an example with several vehicles arriving at different positions and colliding at one point. In this case, reducing the speed will not solve the problem. The vehicles have no priorities compared to each other. Since they are moving in a straight line at the collision point, all three vehicles will be stopped.

\begin{figure}[!h]\centering
\includegraphics[width=0.55\linewidth]{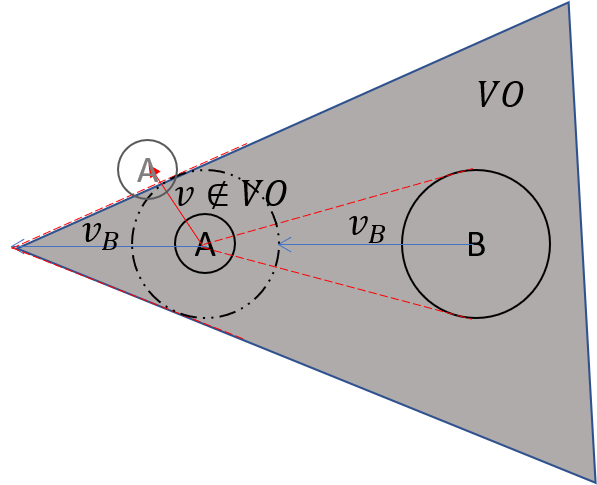}
\caption{Example of determining the velocity vector $v$ of a robot based on the velocity of obstacles. $A$ is the robot, $B$ is a moving obstacle with velocity $v_{B}$. $VO$, the set of velocity obstacle vectors corresponding to obstacle $B$, is shown in the figure. One of the choices of velocity vector for the robot $v\notin VO$ is shown in the figure, but with this choice of $v$, the robot and the obstacle will collide in the next time segment, as shown in the figure.}
\label{fig3}
\end{figure}

\cite{Gyenes:2018} studies the motion planning of multi-robots that ensures the avoidance of static and moving obstacles. In their method, they consider the motion of robots and obstacles in several smaller time periods. They propose an obstacle velocity prediction method that plans the next movement considering the positions and velocities of the robot and obstacles at the current time. They also proposed the safest obstacle velocity method, which not only finds the highest velocity in each time segment, but also finds the safest plan. This is because if only the fastest motion plan is considered, the robot may approach very close to the obstacles, which increases the collision risk due to the inaccuracy of the size, position, and velocity information. The velocity vector decision is made by choosing a velocity vector outside the set of velocity vectors of all moving obstacles, including the boundary information of all obstacles and vectors towards static obstacles and their boundaries. For this purpose, they consider the set of vectors called velocity-obstacle vector set starting from the position of the robot and choose the velocity vector outside this set. In their study, the prediction of the obstacle motion should be known in advance. The set of velocity obstacle vectors starting from the position of the robot does not work for the case when an obstacle is moving towards the robot. They have to include the set of vectors in the direction of the obstacle considering its expected position in the next time segment. Also, they have not considered the size of the robot. An example of this problem can be found in Figure \ref{fig3}.

\cite{Chu:2018} proposed polynomial interpolation for trajectory planning from initial to final state was used to find a continuous trajectory from initial to final state for each motion. To avoid obstacles, they decompose the trajectory from the initial to the final state into several shorter trajectory segments that do not intersect with obstacles, and use polynomial interpolation (polynomials of high degree or splines) to find smooth motions. The given points of the segments should be far away from obstacles to avoid collisions with obstacles at the endpoints. However, the polynomial path may intersect with obstacles regardless of how far the points are chosen. To find the points, we could select the points using Dijkstra's algorithm by considering the workspace as a grid.

In \cite{Yao:2020} the workspace is divided into grids. And the occupancy spaces of robots are identified as $0$ and $1$ values in their three-dimensional matrix representations. The movement of robots leads to changes in the values of their corresponding matrices, and the new matrices are compared to find possible collisions. Their method depends on the size of the grid. If the grid size is large, the current states of the robots can be considered as collisions, but if the grid size is smaller, they cannot.


\cite{Park:2020} studied the collision probability of two objects with uncertainty in their positions. They represented the uncertainty in the positions of the objects with non-Gaussian forms. They claimed that their proposed collision probabilities represent tight bounds on the convex shape (the safe region around the objects), and they used this for motion planning. To deal with the noise due to uncertainties, they used sensory information. The uncertainties of the objects are due to their geometric representation. To do this, they determine the probability distributions of the two objects and find the convolution distribution when one of the objects shifts. This creates the probability regions around the objects, which help to avoid collisions. They find the region in the Minkowski addition of two regions around the objects that have zero probability in the convolution distribution. They considered the truncated Gaussian mixture models for errors that increase the region around the objects and expand the collision probabilities from it.

\cite{Chen:2021a} describes a method for a path planning algorithm to coordinate neighbouring robots and avoid collisions. The problem is translated into a mixed observed Markov decision process and an optimization problem is extracted. Two robots are adjacent if they are in such states that they can reach a fixed state by certain actions. To avoid collisions, actions that cause the robots to reach the same state should be avoided. In this case, robots with possible collisions will perform actions that do not put them in the same state, and the total reward together with the total negative cost are maximal.

\cite{Behrens:2020} described an optimization problem for assigning tasks to robots in a sequence of tasks and motions of robots such that there are no collisions between robots, the predefined set of constraints is satisfied, and overall makespan is minimized. In their method, tasks are divided into confined tasks (a task where a robot's action is limited to a small part of the workspace, e.g., grasping and placing) and extended tasks (a task where a robot's action is limited to a large part of the workspace, e.g., welding along a line). The main focus is on the optimization of the extended tasks. For a given set of tasks that satisfy the set of constraints during execution, each task has multiple starting positions (called degrees of freedom), and the time interval in which a robot can perform each task is measured. The workspace is partitioned to identify regions that a robot occupies during the execution of a task in a sequence of time frames by discretizing the time interval of the task into smaller successive intervals. For solving tasks and motion planning for extended tasks, the problem is translated into a constraint satisfaction problem, which is a type of optimization problem modeled with triples $(X,D,C)$, where $X$ is a set of variables, $D$ is a set of domains where parameters take values, and $C$ is a set of constraints. The solution is to assign values from $D$ to the variables $X$ such that the set of all constraints $C$ is satisfied. To solve the problem, the gradient method and steepest descent by Cauchy \cite{Goldstein:1962} is used (backtracking search method). The optimization model is viewed from three perspectives: task layer, robot layer, and collision-free plan. The time intervals considered in the paper depend only on the task and the dependence on the robots is not considered, i.e., the robots should be identical. Moreover, the size of the region and the discretization of the time interval are not fixed and can be either a short or a long interval. Depending on the choice of the interval size and the size of the domain, different solutions can be obtained. Moreover, in the backtracking search method, in order to obtain a solution for the first upper bound, a random selection seems to be made in the solution space, so that the values in the domains that satisfy the constraints are selected. But the steps to reduce the upper bounds are not described. And the solution completely depends on the selection and reduction of the upper bounds. Moreover, different upper bounds may lead to different solutions. Also, the robot dependency is skipped but should be considered since the navigation codes, time intervals, constraints, and active components are robot-dependent, and their values may change when switching from one robot to another. In \cite{Behrens:2020}, tasks are translated into ordered visit constraints originally defined in \cite{Behrens:2019}. Here, the tasks are considered confined, so the start and end locations are considered identical, and the robot configuration is the same at the start and end. However, the new modified version includes different locations and different configurations to include extended tasks. Moreover, collisions may occur between components of a single robot, which is not considered. In addition, collision avoidance depends on the size of the regions (voxelization sizes). When the region size is large, the robots can have intersections in the configuration spaces, but when the region size is smaller, they have empty intersections. This is consistent with the well-known result that there are always infinitely many other real numbers between two distinct real numbers, see \cite{Gaughan:1993}. This means that if the two robots are not connected at any point, there will always be a pixel size where the intersection of their voxalizations is empty.

\begin{figure}[!h]\centering
\includegraphics[width=0.9\linewidth]{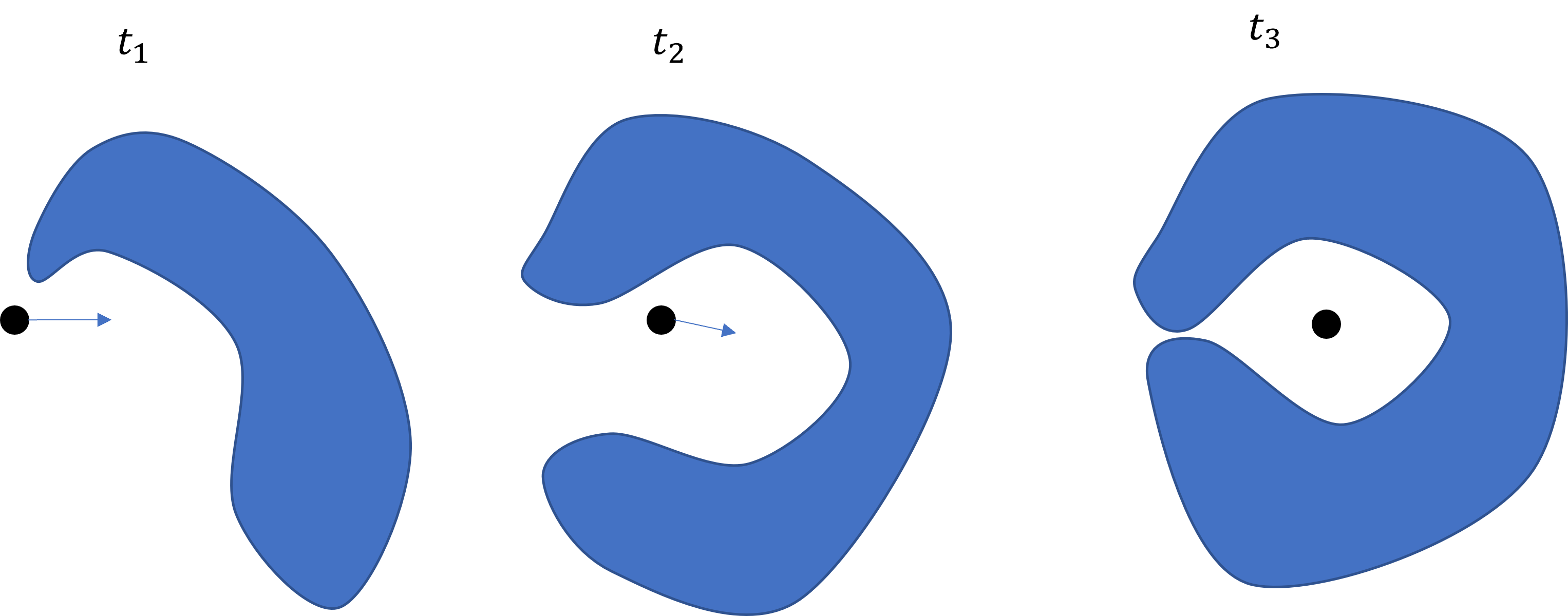}
\caption{Example of a robot surrounded by a deforming object. The deformation appears with time from left to right. After time $t_3$, the robot can no longer avoid the collision.}
\label{fig2}
\end{figure}

\cite{Sunkara:2019} studied the collision avoidance of an object with arbitrary shape and a deforming object, and proposed a nonlinear model for collision avoidance. In this study, the robot and the obstacle can have arbitrary shapes. It needs a constant velocity when the shape of the obstacle changes. If the deforming object has acceleration in a certain direction, collisions may occur because the guidance method does not consider the acceleration of the deforming object. In the proposed method, the boundary of the object is described by descritization and it is assumed that in each segment of the boundary, all points have the same velocity. The method uses Lyapunov function to guide the robot to avoid collision with the deformed object. Thus, to avoid collisions, the system is assumed to be locally Lyapunov-stable \cite{Lyapunov:1992} and the velocity vector of the robot is always correlated with the velocity of the deformed objects according to the collision avoidance guidance. Since the formulation considers the local region of the object near the robot, there are some examples where the robot is surrounded by the deforming object without the possibility of avoiding a collision, see Figure \ref{fig2}. Thus, without knowing the global pattern of the deforming object, the local guidance can also cause a collision.

\cite{Wing:2020} studied motion planning and collision avoidance without predicting the velocity of moving obstacles. In this method, the authors consider a safe distance around the obstacle and take the velocities of the obstacle and the robot to determine the angular velocity of the robot to avoid a collision. The method is a geometric approach that maps the motions of all objects in 2D space. To avoid a collision, the direction of the robot's velocity changes to an angle that is either opposite to the direction of the obstacle's velocity (when the obstacle moves in front of the robot) or to the direction that corresponds to the edge of the obstacle and has the shortest distance from the line generated by the robot's original direction (when the obstacle moves toward the robot). In the proposed method, the velocity of the obstacles should be measured at all times and sudden changes in the direction of the obstacles are not considered. Since the safety distance around all objects is considered, the method does not focus on the minimum total distance. Also, the method should consider the size of the robot, since larger robots may need a larger change in angular velocity than smaller robots to avoid collisions.

\cite{Zhang:2021} uses a topological approach to cover all objects with convex sets. In this way, the collision avoidance motion becomes a smooth function instead of a discrete function. The method penalizes the generated trajectory, which helps to find the least disturbing trajectory when a collision cannot be avoided. The method uses the property of the mathematical concept of compact set, which is defined as a set where each cover has a finite subcover, \cite{willard:2004}. This concept is purely theoretical and there is no unique way to find finite subcovers. Even for a given cover, the set of finite subcovers cannot be uniquely identified. After finding a finite cover, the union of all elements forms a bounding box around the robot. However, such a finite cover does not mean that the area covered by the union is minimal.

\cite{Hajiloo:2021} studied the threshold for direction change, braking, and acceleration of an autonomous vehicle with the minimum distance to obstacles that suddenly appear in the path of the vehicle, so that the vehicle remains stable. In this study, the friction of the road is considered as an important parameter that contributes to the stability of the vehicle motion and the controller handles the motion, angle and braking of all wheels. They translate the motion into a dynamical system and solve the optimal planning of the motion of all wheels in case an obstacle suddenly appears by changing the acceleration of each wheel.

\begin{figure}[!h]\centering
\includegraphics[width=0.8\linewidth]{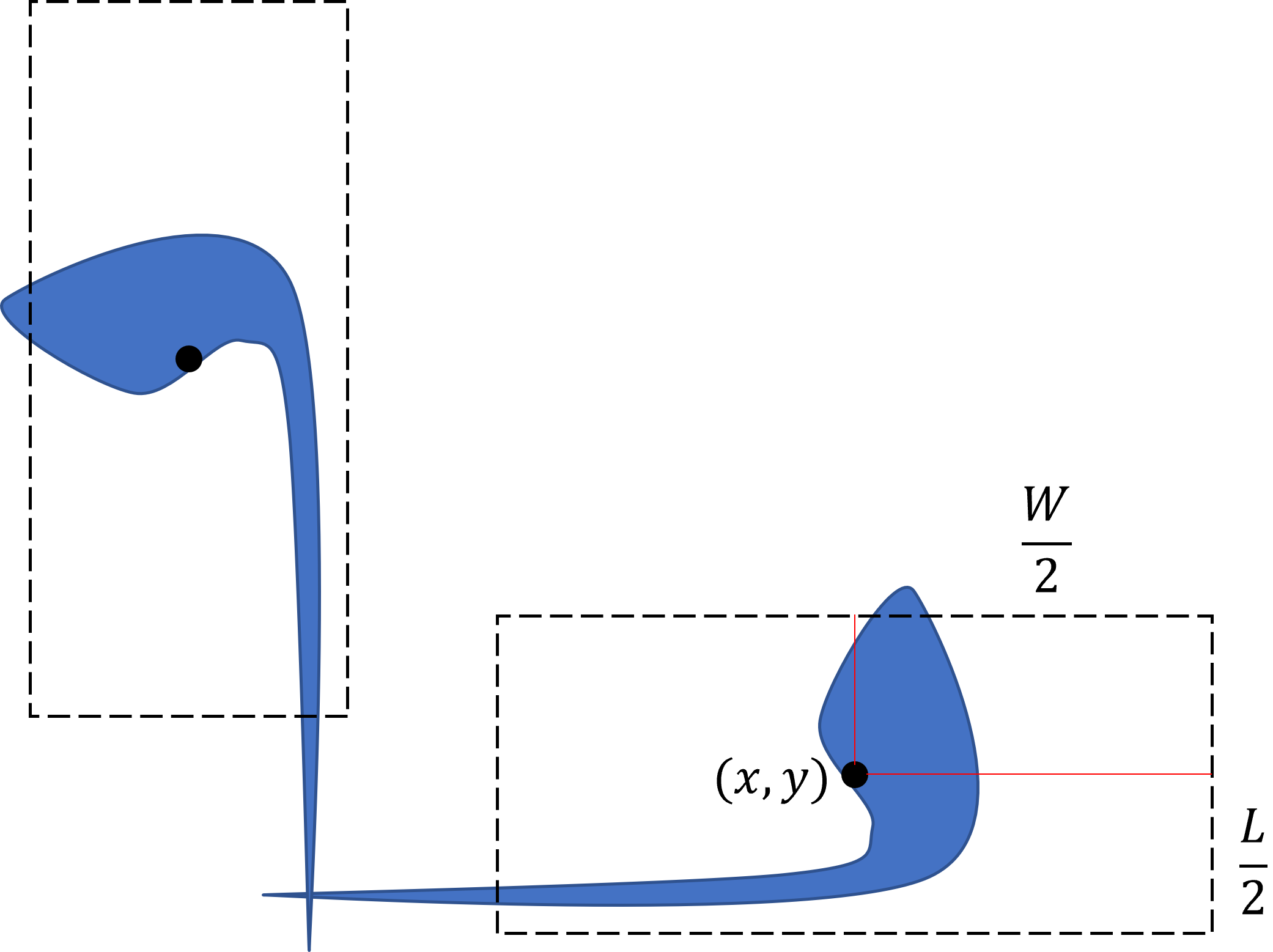}
\caption{Example of irregularly shaped vehicles. The rectangular area created does not completely cover the vehicle. And there is a possibility of collision between the vehicles.}
\label{fig4}
\end{figure}

In \cite{Nakamura:2020}, the authors assume that all vehicles send their local status and planned trajectory to a server, and the server determines whether there is a possible collision in a group of vehicles within a certain time. Then, the server sends a modified trajectory of the vehicles to avoid collisions. Their proposed method is a centralized trajectory planning method that reduces the computation time for collision detection and avoidance by removing vehicles without possible collisions. They assume that each vehicle has a rectangular area around it and consider a collision when the rectangular area around a vehicle intersects with the road boundary, obstacles, and the rectangular areas of other vehicles. And to avoid collision, they consider the change of acceleration vector of vehicles in case of possible collision. The proposed method considers the rules of passing a road, distinguishes the case of intersections and roundabouts, and treats them independently. The method of finding a rectangular box around a vehicle can only be applied to vehicles with regular shape and cannot be generalized to all vehicles with irregular shape. In this method, they find the center of gravity of the vehicle and create a rectangle by adding half the sizes of the length and width of the vehicle shape and assume that the resulting rectangle surrounds the vehicle. Figure \ref{fig4} shows an example of when the created rectangular area does not surround the vehicle. In these cases, part of the vehicle may be outside the rectangular area, which may intersect with other vehicles but is not considered in the collision avoidance.

\begin{figure}[!h]\centering
\includegraphics[width=1\linewidth]{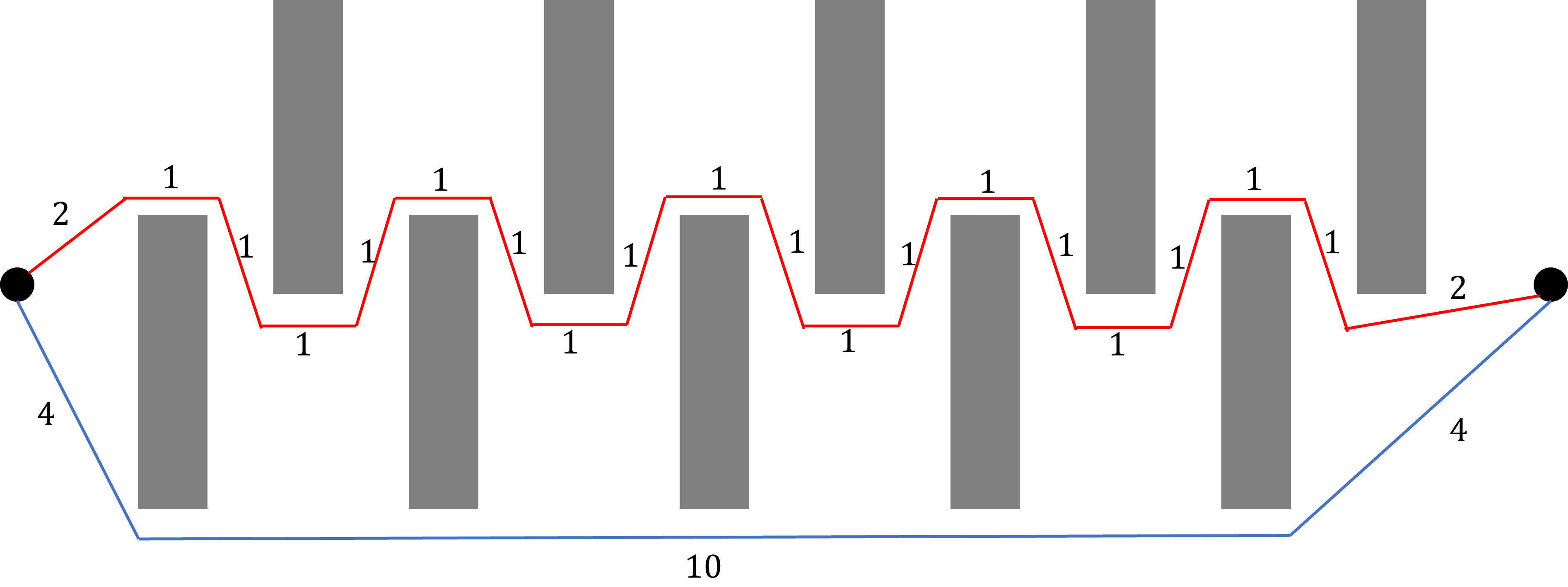}
\caption{The shortest path cannot be determined by random sampling near the obstacle region. The shortest path (in blue) is 18, but the path by random sampling (in red) is 22.}
\label{fig5}
\end{figure}

\cite{Rakita:2021} studied path planning with collision avoidance using combined graph theoretic and probabilistic approaches. In the proposed method, the workspace is divided into an obstacle space and an obstacle-free space. Then, first, the initial position is connected to the final position by a straight line. When the straight line intersects with the obstacle space, points near the first obstacle are randomly selected and a tree is created connecting the initial point with the selected points. Then select a point from the sample where the straight line from its position to the final point does not intersect the first obstacle, and select it as the first position of the trajectory from the initial point to the final point. This method only works if the obstacle space is completely known and all obstacles are static. If the robot has to explore the environment to detect obstacles and if the obstacles can move, the method does not work. The method also does not consider minimising the total distance the robot travels. See Figure \ref{fig5} for an example of when the shortest path that avoids collisions is longer than the path determined using random sampling. 

\cite{Hai:2021} studied motion planning and collision avoidance for multi-robot in dynamic environment. The proposed method attempts to answer the question of what is the collision-free trajectory of each robot when the future goal of the robots is known. The proposed method is a decentralized model where the prediction of robot motion and behavior is learned through demonstration using a centralized sequential planner (based on recurrent neural networks (RNNs)). In the proposed model, by design, there is no communication between the robots and uncertainties in the motions of obstacles and robots are not considered.

\begin{figure}[!h]\centering
\includegraphics[width=1\linewidth]{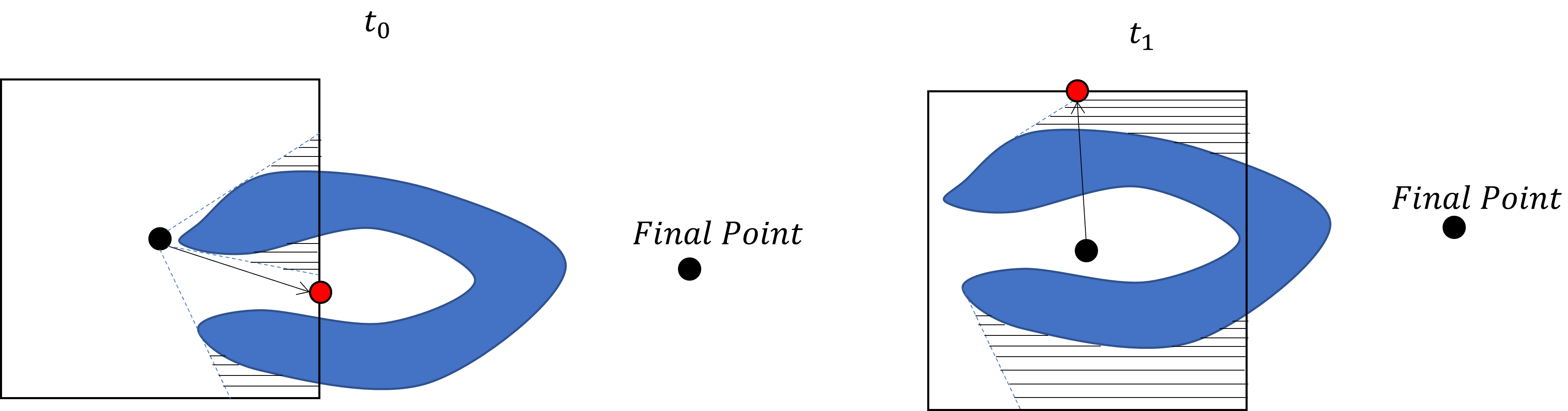}
\caption{The robot moves to the points with the highest weights in two time segments and in the last time segment the movement to the point with the highest weight collides with the obstacle. The dashed area is the area that is not explored for possible collisions. The red circles are the points with the highest weights.}
\label{f2}
\includegraphics[width=0.5\linewidth]{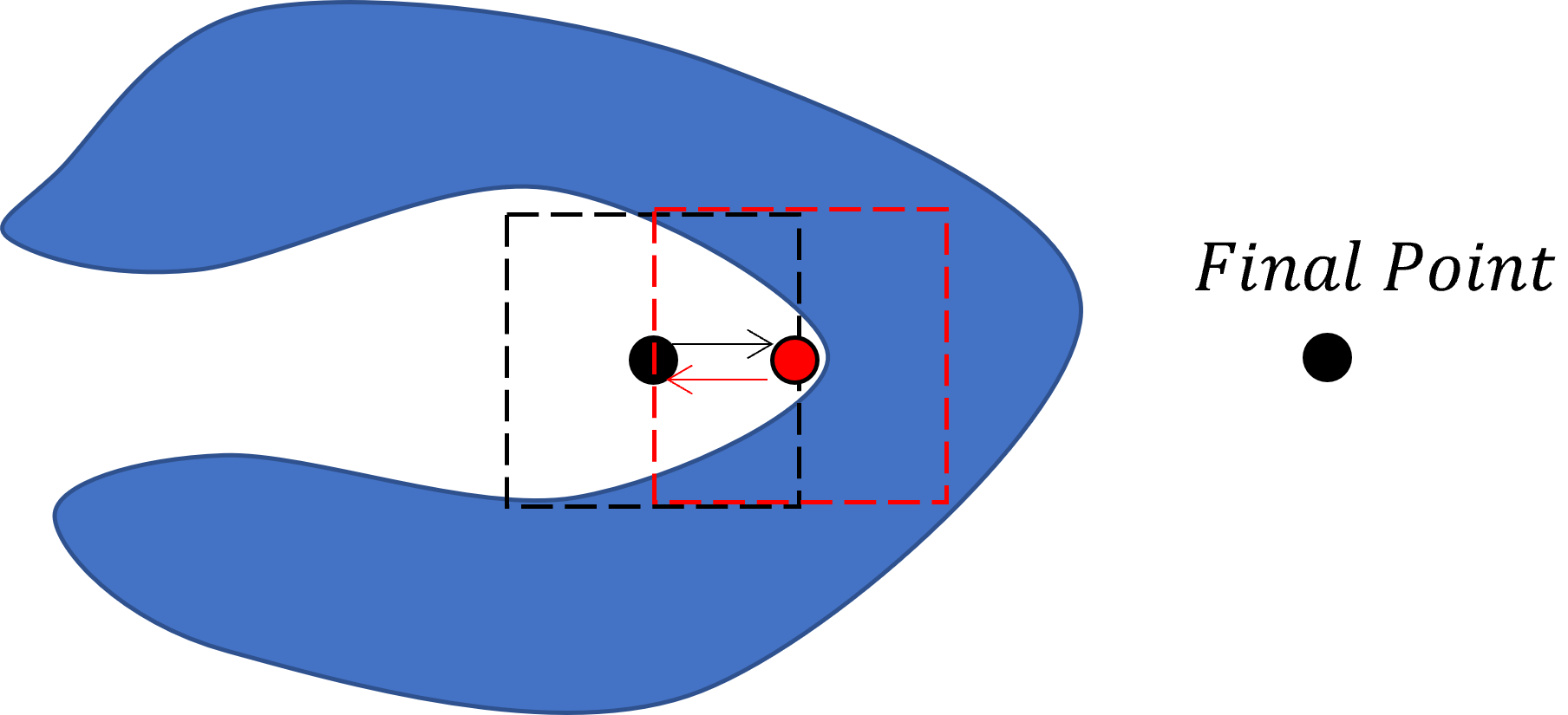}
\caption{The robot moves infinitely back and forth to the point with the highest weight.}
\label{f3}
\end{figure}

The paper \cite{Fox:1997} described a simple geometrical model of the motions. The authors studied motion planning and collision avoidance by minimizing the travel distance and maximizing the speed of the robot. Their method uses what they call dynamic windows, which is a set of velocities that the robot can reach within a short period of time, allowing it to safely reduce or accelerate when it detects a potential collision. In this method, the robot's velocity is considered to be a constant value in each time segment, and it is assumed that the robot is the center of a circle whose radius is equal to the distance that the robot can travel at its current velocity (in \cite{Fox:1997}, they immediately switched from circle to rectangle). To model the robot's motion, angular acceleration was considered to allow the robot to change direction. Then, the velocity and angular acceleration are approximated while the position of the robot is updated using the information obtained from its wheels. In each time segment, the robot finds the rectangular area around the robot in which it can move. The area is obtained from the maximum speed that the robot can reach considering its predicted velocity and the angular acceleration of the robot. Then, each point of the obtained rectangular area is assigned three weights for the distance to an obstacle, the distance to the goal, and the velocity of the robot to reach that point, where the weight for the distance to an obstacle is higher for points farther from the obstacle, higher for points closer to the goal, and higher for points farther from the robot. The main optimization problem given in \cite{Fox:1997} is 
$$G(v,w)=\sigma(\alpha.heading(v,w)+\beta.dist(v,w)+\gamma.vel(v,w)),$$
where $v$ is the velocity at which the robot moves straight, $w$ is the angular velocity at which the robot changes direction, $heading$ is the distance from the robot's next position to its final position, $dist$ is the distance from the robot's next position to the obstacle, and $vel$ is the speed at which the robot moves to its next position. The movement to the selected point with the highest weights may collide with an obstacle, see Figure \ref{f2}. Even if they ignore the history of the area under study and include only the points with the highest weights, their proposed method may enter an infinite loop of moving back and forth, for example, in Figure \ref{f3} and the second time segment, the optimal point to move to without considering history is the point closest to the obstacle, while the next time segment is the point where the robot was. It is also assumed that the obstacles are static.

\begin{figure}[!h]\centering
\includegraphics[width=0.5\linewidth]{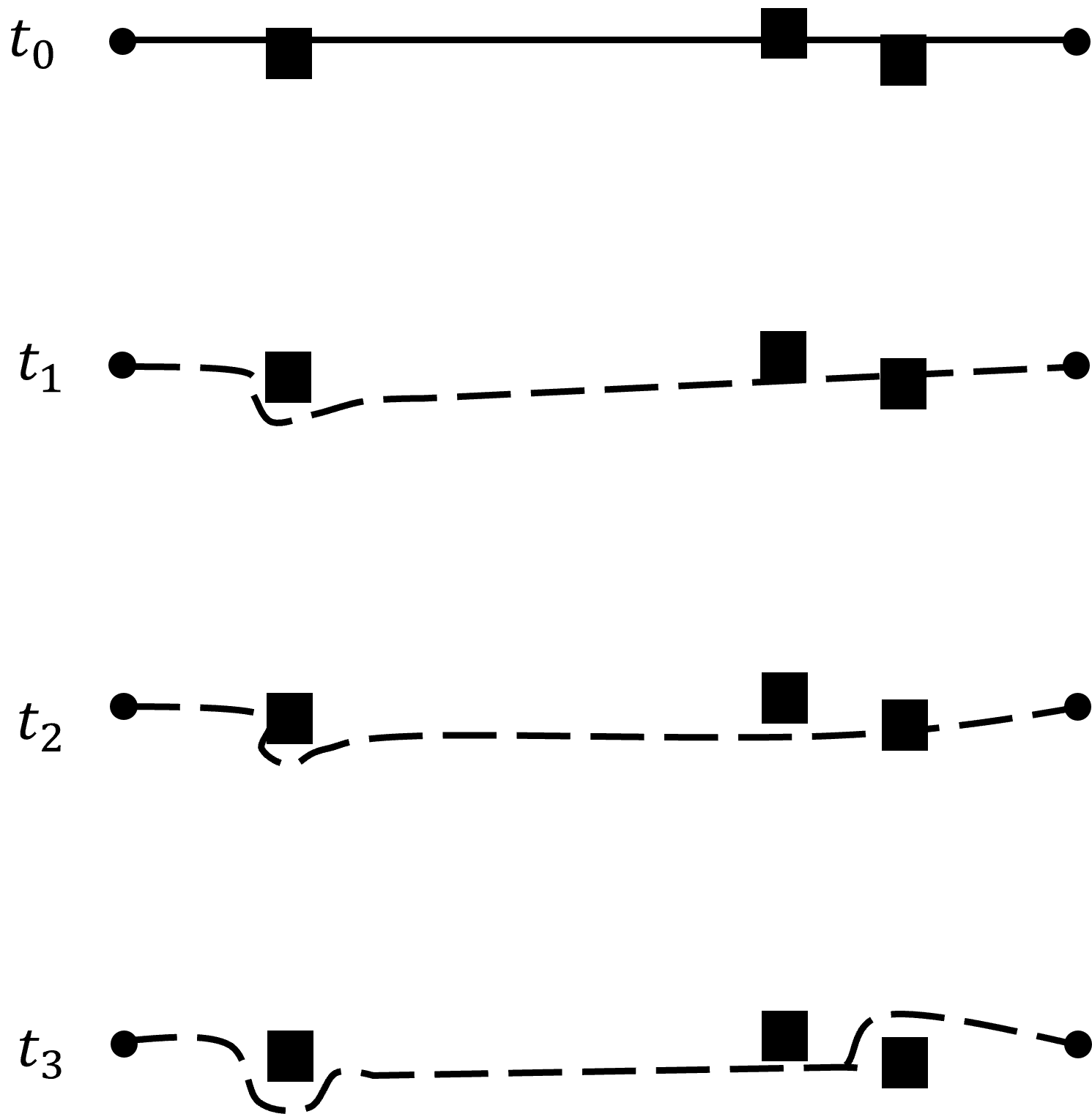}
\caption{Intuition for elastic band. Starting from a straight line, at each time step where an obstacle is observed, a section of the line is replaced by a smooth curve.}
\label{f4}
\end{figure}

\cite{Rosmann:2012} and \cite{Rosmann:2013} have studied motion planning with collision avoidance. In these methods, the path from the current position of the robot to the final point is considered as an elastic band, where after detecting each obstacle, the path is replaced by a curve from the current position of the robot to the final point that does not intersect with the obstacle, see Figure \ref{f4}. This method helps to improve the method in \cite{Fox:1997} by incorporating history (since the information about the band is stored in each time segment), avoiding an infinite loop, and avoiding scenarios where the robot moves through an obstacle by moving straight forward, see Figure \ref{f2}. The method in \cite{Rosmann:2012} does not optimize the robot's velocity along the path, and this problem is solved in \cite{Rosmann:2013}. The authors in \cite{Rosmann:2013} create a graph for the curved path by using the positions on the curved path as nodes and obtaining the edges by the sequential order of the positions on the path, taking into account the time differences in moving from one position to another. This time helps to change the acceleration of the robot and control the translational and angular velocity of the robot to further minimize the time to reach the final point. Since the methods in \cite{Rosmann:2012} and \cite{Rosmann:2013} use curves instead of direct lines, the magnitude of the distance traveled may not be minimal. Also, these methods do not account for moving obstacles.
If the time steps are very short and the distance to the final position is long and there are many obstacles, the method requires a lot of memory to store the shape of the curve (all points on the curve).

The work \cite{Rosmann:2020} deals with motion planning with obstacle avoidance. They combine the Euclidean measure with a rotational component for motion planning and collision avoidance, where the goal is to minimize time. They model the trajectory as a function of time as a continuous nonlinear differential equation with bounded derivatives at each time point. Their goal is to minimize the cost of moving from the starting point to the end point. Since the robot can change direction (a rotation is applied) when moving to the next state (the next time step), the new operation should be defined to handle this non-Euclidean motion. For this purpose, in $2D$-space, they consider the motion at each time step as an operator in the special orthogonal group ($SO(2)$), which means that the velocity vector of the robot changes its direction with an angle. Since the rotation operation is only on a circle of radius $1$, they should use the special orthonormal group $\mathbb{S}^1$ instead of $SO(2)$. Later, they normalize the angle that maps the rotation to a circle of radius $1$. One of the constraints is that the dynamic error of the system should be small. In their formulation, the value of the error is considered to be equal to $0$ as one of the constraints. Moreover, for the finite difference kernel between two successive states, they used the implicit second-order Runge-Kutta (Crank-Nicolson) method, which is the average of the values of the state function on the two states. This can be replaced by the higher order implicit Runge-Kutta method to better describe the dynamics of the robot motion. If we use only the second-order difference, the effects of the motions at later time steps in the future are ignored. To avoid collisions, they specify the minimum distance between the robot and an obstacle and feed it as a constraint to find the region in which the robot can move without collisions while maintaining the specified minimum distance to obstacles. This is equivalent to considering a bounding box around all obstacles.

\section{Model}
The idea is to plan the motion of a robot in a dynamic environment with both deterministic and non-deterministic objects, creating a safe zone around all objects to avoid collisions, predicting the motions of non-deterministic objects, and taking into account the dimension of the objects. The main idea is to discretize all motions, define safe zones around all objects based on velocity vector probabilities, and find an optimal direction and velocity of the robot at each time step with the lowest cost (minimum travel distance at minimum time). To formulate the model, we use the following:
\begin{itemize}
\item\textbf{Object}: It is the set of all robots, humans, and obstacles that either change their position at a given time step or maintain their initial position at all times. Objects are identified as $i\in\{1,\ldots,N\}$.
\item\textbf{Map}: It is the environment $Env$ where all objects and the robot are located, and their movements are observed or predicted.
\item\textbf{Robot velocity limit}: It is the maximum velocity value of the robot, denoted by $\mathrm{v}$.
\item\textbf{State of an object}: For object $i$, at time step $t$, it is a tuple 
\begin{equation}\label{eq:eq1}
(orien_t^i, dist_t^i,d_{v^i_t},P(d_{v^i_t},\-),v^i_t,P(v^i_t,\-)),
\end{equation}
where 
$$(orien_t^i, dist_t^i)\in[0,2\pi)\times\mathbb{R}_{\geq0}$$ 
are the polar coordinates of the object $i$ at time $t$, $v^i_t\in\mathbb{R}_{\geq0}$ is the expected velocity value, $d_{v^i_t}\in[0,2\pi)$ is the expected direction of the object, $P(d_{v^i_t},\-)$ is the probability density function of the direction error of the object $i$ at time step $t$, and $P(v^i_t,\-)$ is the probability density function of the error of the velocity value of the object $i$ at time $t$.
\item\textbf{Deterministic object}: It is such an object whose state at each time step $t$, is 
\begin{equation}\label{eq:eq2}
(orien_t^i, dist_t^i,d_{v^i_t},1,v^i_t,1).
\end{equation}
Obstacles and objects whose motions are known at each time step are deterministic objects.
\item\textbf{Non-deterministic object}: It is an object whose state at a given time step $t$, is given by \eqref{eq:eq1} with either $P(v^i_t,\-)\neq1$ or $P(d_{v^i_t},\-)\neq1$.
\item\textbf{Probability density function $P(v^i_t,\-)$}: Since for the cases where the expected velocity value is $v^i_t > 0$, the actual velocity should be $v^i_t+\Delta v$, this means that the object $i$ has a slightly higher or lower velocity than the expected velocity value ($\Delta v$ as an error). To increase the accuracy of the expected (predicted) velocity, the $\Delta v$ value should be close to $0$, with the same probabilities of higher or lower velocity. Now, if $v^i_t=0$, $\Delta v\geq0$. Therefore, $\Delta v$ should be considered as a truncated normal distribution in the interval $[-v_t^i,\infty)$, with $\mu=0$ and $\sigma_v > 0$. So 
$$P(v^i_t,x)=\frac{\sqrt{2}}{\sqrt{\pi}\sigma_v}\frac{ \exp(-\frac{x^2}{2\sigma_v^2})}{1-\erf(\frac{-v_t^i}{\sigma_v})},$$
where $\erf$ is the Gaussian error function:
$$\erf(x)=\frac{2}{\sqrt{\pi}}\int_{0}^x\exp(-t^2)dt.$$
\item\textbf{Probability density function $P(d_{v^i_t},\-)$}: Since for the cases where the expected direction is $d_{v^i_t}\in[0,2\pi)$, the actual direction should be $d_{v^i_t}+\Delta\theta$, which means that the object $i$ has a slight angular deviation from the expected direction ($\Delta\theta$ as error), where to increase the accuracy of the predicted direction $\Delta\theta$ should be close to $0$, with the same probabilities for clockwise and counter-clockwise angles. If the object $i$ does not move at time step $t-1$, the motion at time step $t$ could be in any direction with uniform distribution $P(d_{v^i_t},x)=\frac{1}{2\pi}$, where $x\in[0,2\pi)$. If the expected direction is $d_{v^i_t}\in[0,2\pi)$, the error parameter $\Delta\theta$ should be a truncated normal distribution in the interval $[-\pi,\pi)$, with $\mu=0$ and $\sigma_d > 0$. So 
$$P(d_{v^i_t},x)=\frac{\sqrt{2}}{\sqrt{\pi}\sigma_d}\frac{\exp(-\frac{x^2}{2\sigma_d^2})}{\erf(\frac{\pi}{\sigma_d})-\erf(\frac{-\pi}{\sigma_d})}.$$
\end{itemize}
The probability density functions $P(d_{v^i_t},\-)$ and $P(v^i_t,\-)$ describe the probability models for the errors in direction and velocity value. The magnitude of the errors in direction and velocity value at time $t$ are independent of the values at all other time steps.
\begin{itemize}
\item\textbf{State of the robot}: For the robot $r$, at time step $t$, it is a tuple 
$$(orien_t^r, dist_t^r,d_{v^r_t},1,v^r_t,1),$$ 
where $(orien_t^i, dist_t^i)$ is the polar coordinate of the robot at time step $t$, $v^r_t\in\mathbb{R}_{\geq0}$ is the decision for the velocity value of the robot, $d_{v^r_t}$ is the decision for the direction of the robot at time step $t$. 

\item\textbf{Safe zone}: It is an area around an object at time step $t$ that the robot cannot visit because of the risk of collision. The velocity value, direction, and size of the object are taken into account.

To find the safe zone for object $i$, at time $t$ in state
$$(orien_t^i, dist_t^i,d_{v^i_t},P(d_{v^i_t},\-),v^i_t,P(v^i_t,\-)),$$ 
find the errors of the angle $\Delta\theta$ and the velocity value $\Delta v$. Find the points $A$ and $B$, where the point $A$ is the last point (sometimes a set of points) of the object, so that when considering parallel lines to 
$$\max\{d_{v_t^i}+\Delta\theta,d_{v_t^i}-\Delta\theta\}$$
and by increasing the width of the origin of the lines, the object and the line intersect, and after slightly increasing the width of the origin of the line, the line and the object do not intersect. And the point $B$ is the last point (sometimes a set of points) of the object, so that when parallel lines are viewed at 
$$\min\{d_{v_t^i}+\Delta\theta,d_{v_t^i}-\Delta\theta\}$$
and by decreasing the width of the origin of the lines, there will be intersections between the object and the line, and after we decrease the width of the origin of the line a little, the line and the object will not intersect, see Figure \ref{lpoint}.

\begin{figure}[!h]\centering
\includegraphics[width=0.8\linewidth]{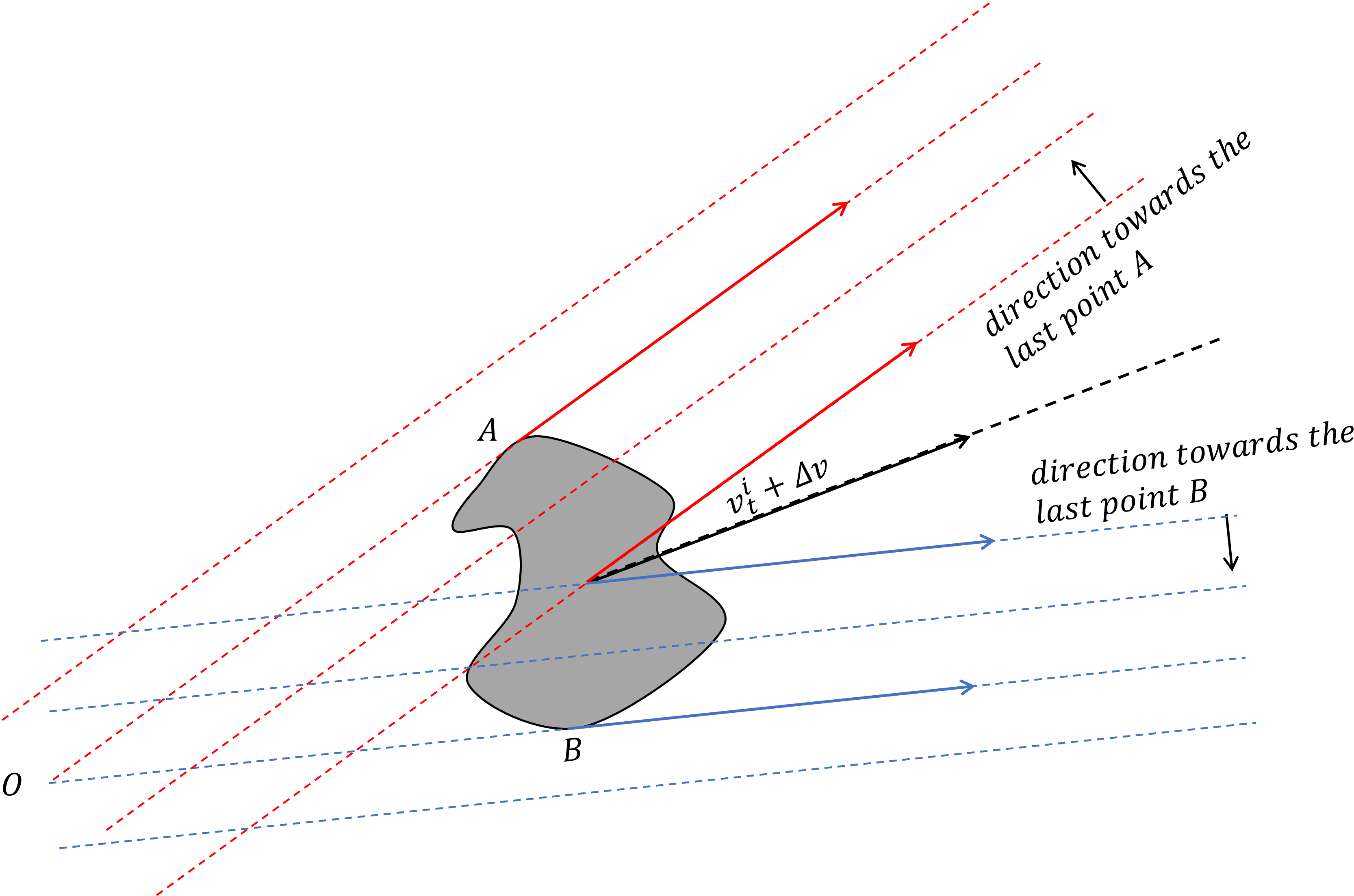}
\caption{Intuition for the method of determining the last points $A$ and $B$, using parallel lines to the directions of the maximum and the minimum expected velocity of the object.}
\label{lpoint}
\end{figure}

Now we should find either the curve $f(A,B)$ (the curve of the object $i$ from $A$ to $B$) and $g(B,A)$ (the curve of the object $i$ from $B$ to $A$) or their respective convex hulls enclosing $f(A,B)$ and $g(B,A)$. Then find the coordinates $A_c$ and $B_c$, respectively, for the motion of the object about the direction 
$$\max\{d_{v_t^i}+\Delta\theta,d_{v_t^i}-\Delta\theta\}~~\text{and} ~~\min\{d_{v_t^i}+\Delta\theta,d_{v_t^i}-\Delta\theta\}$$ 
with scale $v_t^i+\Delta v$. Since we are working with a polar system, $g(B,A)$ is transformed with this scale to be 
$$g_c(B_c,A_c)=(v_t^i+\Delta v)*Map(g(B,A))$$ 
and $f(A,B)$ is transformed to 
$$f_c(A_c,B_c)=(v_t^i+\Delta v)*Map(f(A,B)),$$ 
where for each
$$\alpha\in[\min\{deg(A),deg(B)\},\max\{deg(A),deg(B)\}],$$
with $deg(X)$ is the degree of the point $X$ on the boundary of the object $i$ to the centroid of the object $i$ in $[0,2\pi)$, the mappings $Map(g(B,A))$ and $Map(f(A,B))$ can be found as functions of $\alpha$ as follows:
First find $\alpha_{\max}(h(X,Y))$ and $\alpha_{\min}(h(X,Y))$:
\begin{align*}
\alpha_{\max}(h(X,Y))=&\max\{\beta\mid dist(h(X,Y))\mid_{[\beta,\max\{deg(X),deg(Y)\}],0)} \\
&~~\text{ is strictly decreasing.}\}
\end{align*}
and 
\begin{align*}
\alpha_{\min}(h(X,Y))&=\min\{\beta\mid dist(h(X,Y))\mid_{[\min\{deg(X),deg(Y)\},\beta],0)} \\
&~~\text{ is strictly increasing.}\}.
\end{align*}
Then we define for $Z\in h(X,Y)$:
\begin{align*}
&\Gamma_{1,\max}(Z,h(X,Y))=\\
&\left\{\begin{array}{ll}
\max\{h(X,Y)\mid_{\beta}\mid\beta\in[\alpha_{\min}(h(X,Y)),deg(Z)]\},&deg(Z)\geq\alpha_{\min}(h(X,Y))\\
h(X,Y),&otherwise,
\end{array}\right.&
\end{align*}
\begin{align*}
&\Gamma_{2,\max}(Z,h(X,Y))=\\
&\left\{\begin{array}{ll}
\max\{h(X,Y)\mid_{\beta}\mid\beta\in[deg(Z),\alpha_{\max}(h(X,Y))]\},&deg(Z)\leq\alpha_{\max}(h(X,Y))\\
h(X,Y),&otherwise,
\end{array}\right.
\end{align*}
\begin{align*}
&\Gamma_{1,\min}(Z,h(X,Y))=\\
&\left\{\begin{array}{ll}
\min\{h(X,Y)\mid_{\beta}\mid\beta\in[\alpha_{\min}(h(X,Y)),deg(Z)]\},&deg(Z)\geq\alpha_{\min}(h(X,Y))\\
h(X,Y),&otherwise,
\end{array}\right.
\end{align*}
and
\begin{align*}
&\Gamma_{2,\min}(Z,h(X,Y))=\\
&\left\{\begin{array}{ll}
\min\{h(X,Y)\mid_{\beta}\mid\beta\in[deg(Z),\alpha_{\max}(h(X,Y))]\},&deg(Z)\leq\alpha_{\max}(h(X,Y))\\
h(X,Y),&otherwise.
\end{array}\right.
\end{align*}
Then
$$Map(g(B,A))\mid_{\alpha}=\max\{\Gamma_{1,\max}(Z,g(B,A)),\Gamma_{2,\max}(Z,g(B,A))\mid Z\in g(B,A)\}$$
and 
$$Map(f(A,B))\mid_{\alpha}=\min\{\Gamma_{1,\min}(Z,f(A,B)),\Gamma_{2,\min}(Z,f(A,B))\mid Z\in f(A,B)\}.$$
Intuitively, the maps $Map(g(B,A))$ and $Map(f(A,B))$ can be viewed as repeating local maxima and local minima within the interval of the first and last local maxima and the first and last local minima, respectively. Note that, by construction, the degree of points $A$ and $B$ with respect to the centroid of the object $i$ are the degrees between
$$\max\{d_{v_t^i}+\Delta\theta,d_{v_t^i}-\Delta\theta\}~~\text{and} ~~\min\{d_{v_t^i}+\Delta\theta,d_{v_t^i}-\Delta\theta\},$$
each with respect to the origin. So any point $X=(r,x)$, where $r$ is the radius to the centroid and $x$ is the angular degree, can be viewed as the vector $(r\cos(x),r\sin(x))$ with respect to the centroid in the Euclidean metric. Also, the current position of the centroid of the object $(R,\beta)$ can be viewed as the vector $(R\cos(\beta),R\sin(\beta))$ with respect to the origin in the Euclidean metric. This means that the coordinate of the point $X$ with respect to the origin in the Euclidean metric is the sum of two vectors:
\begin{align*}
(r\cos(x),r\sin(x))&+(R\cos(\beta),R\sin(\beta))=\\
&(r\cos(x)+R\cos(\beta),r\sin(x)+R\sin(\beta)),
\end{align*}
which can be easily converted into a polar metric as the point $Y=(\mathcal{R},\mathcal{\gamma})$, where 
$$\mathcal{R}=\sqrt{(r\cos(x)+R\cos(\beta))^2+(r\sin(x)+R\sin(\beta))^2}$$
and
\begin{align*}
\gamma=&\tan^{-1}\left(\frac{r\sin(x)+R\sin(\beta)}{r\cos(x)+R\cos(\beta)}\right),\\
&\text{with}~~\gamma\in{[\min\{d_{v_t^i}+\Delta\theta,d_{v_t^i}-\Delta\theta\},\max\{d_{v_t^i}+\Delta\theta,d_{v_t^i}-\Delta\theta\}]}.
\end{align*}
So, 
$$Map(X)=Y.$$ 
Note that the initial centroid has coordinates $(R,\beta)$ with respect to the origin, the new centroid of the object for the predicted position has coordinates $Predict\_Centroid=(v_t^i+\varepsilon,d_{v_t^i}+\delta)$ with respect to the initial centroid. Now suppose that $X$ is a point on the curve $f(A,B)$ (or on the curve $g(B,A)$), then its corresponding point $New\_X$ on the curve $f_c(A_c,B_c)$ (or on the curve $g_c(B_c,A_c)$) is obtained as follows: 
\begin{align*}
New\_X&=Polar(Euclidean(Map(X),O)\\
&~~~~+Euclidean(Predict\_Centroid,(\mathcal{R},\gamma)))\\
&=Polar(Euclidean((\mathcal{R},\gamma),O)+Euclidean(X,(\mathcal{R},\gamma))\\
&~~~~+Euclidean(Predict\_Centroid,(\mathcal{R},\gamma)))
\end{align*}
which is a polar transformation of the sum of the vectors of the Euclidean coordinates of the points $X$ with respect to the origin and $Predict\_Centroid$ with respect to the initial centroid coordinate. See Figure \ref{figg1}.

\begin{figure}[!h]\centering
\includegraphics[width=0.5\linewidth]{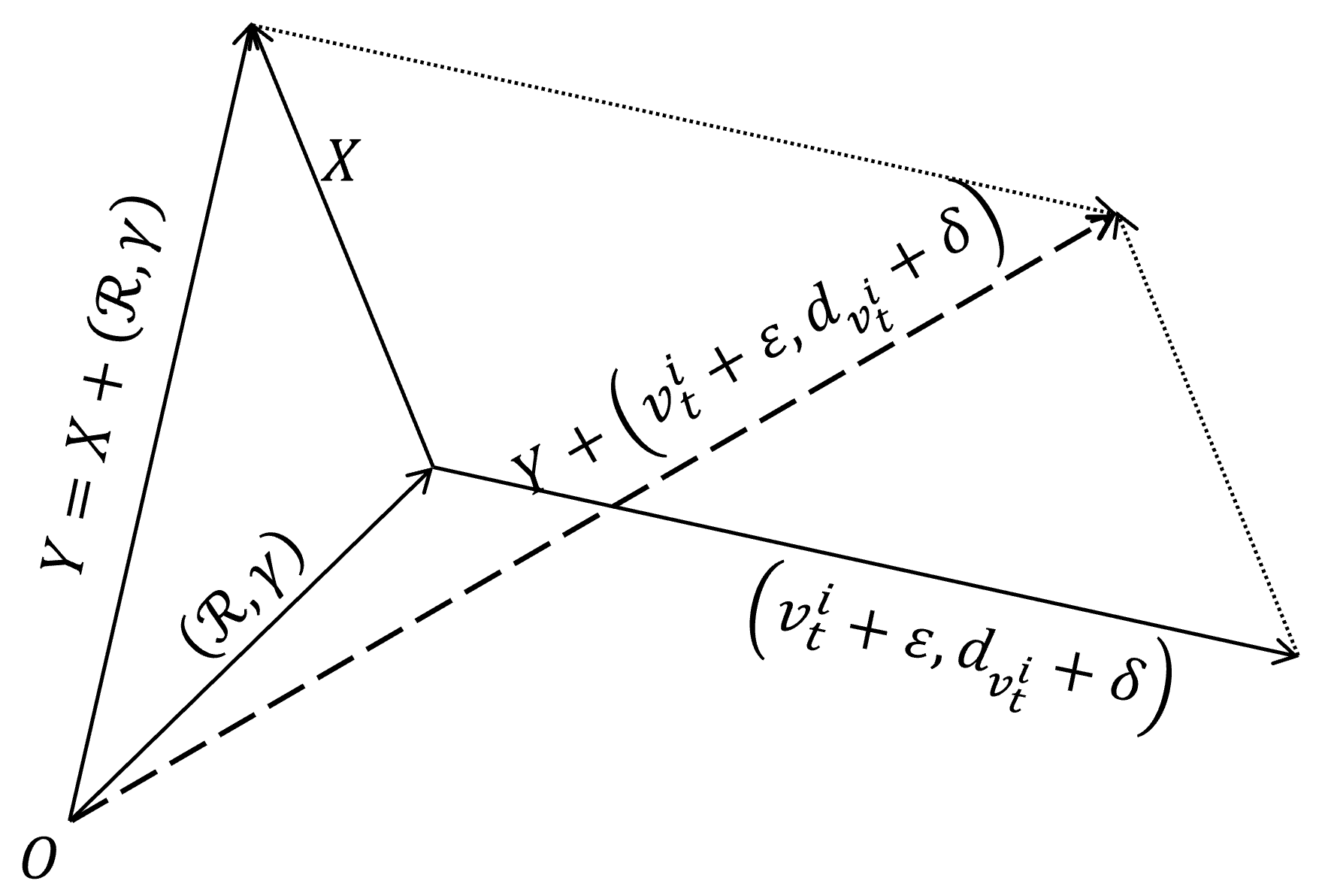}
\caption{Intuition for the method of determining the coordinates of each point of an object after predicting its motion. The dashed vector is the coordinate of interest, i.e., the coordinate of the point $X$ of the object $i$ after applying the predicted motion of the object. Dotted vectors are used to indicate parallel vectors.}
\label{figg1}
\end{figure}

Now, the safe zone of object $i$ at time step $t$ with state given by \eqref{eq:eq1} can be obtained by 
$$SZ^i_{t+1}=\max\left\{\int_{d_{v^i_t}-\Delta\theta}^{d_{v^i_t}+\Delta\theta}(g_c(B_c,A_c)^2-f(A,B)^2),\int_{d_{v^i_t}-\Delta\theta}^{d_{v^i_t}+\Delta\theta}(f_c(A,B)^2-g(B,A)^2)\right\}.$$

See Figure \ref{figg2} for the intuition of obtaining the safe zone.
\begin{figure}[!h]\centering
\includegraphics[width=0.8\linewidth]{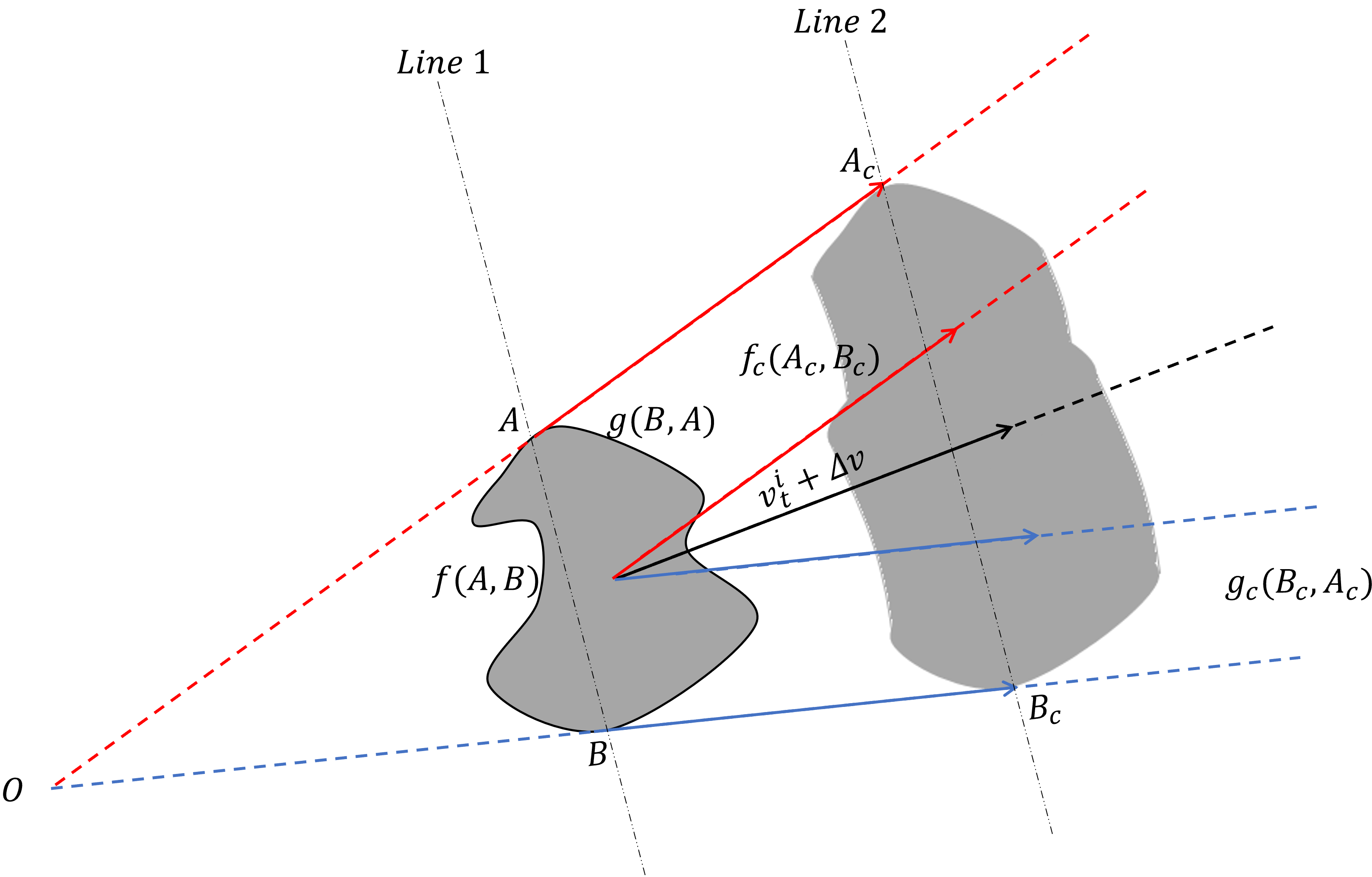}
\caption{Intuition for the method of obtaining the safe zone. $g_c(B_c,A_c)$ ($g(B,A)$) is the boundary curve of the map of the object (the object) from $B_c$ ($B$) to $A_c$ ($A$), that is, the boundary curve of the map of the object (the object) to the right of Line 2 (Line 1). $f_c(A_c,B_c)$ is the boundary curve of the map of the object (the object) from $A_c$ ($A$) to $B_c$ ($B$), i.e., the boundary curve of the map of the object (the object) to the left of Line 2 (Line 1).}
\label{figg2}
\end{figure}

\item\textbf{The optimization objective}: Let the robot aims to move from point $A$ to point $B$. Let $Shortest\_Path (A,B)$ be the shortest path from $A$ to $B$ without considering moving objects and $|Shortest\_Path (A,B)|$ be the total distance of the shortest path and assume that the robot can travel the shortest path in time $t$. Let $act_t$ be the actual position of the robot at time step $t$. The actual position of the robot is a function of the initial position, the direction of the robot, and the velocity value of the robot. The positions are denoted as a sequence $P_0=A,P_1,\ldots,P_{T-1},P_{T}=B$. The objective is then to minimize the following 
$$\min_{P_0=A,P_1,\ldots,P_{T-1},P_{T}=B~\text{ valid }}\sqrt{\left(\frac{T}{\mathrm{t}}\right)^2+\left(\frac{\left(\sum_{t=0}^T(\min_{P\in Shortest\_Path(A,B)}\{(P_t-P)^2\})\right)}{|Shortest\_Path(A,B)|^2}\right)^2},$$
where a path is considered valid if there are no collisions with other objects. Note that the time and distance are divided by the minimum time and length of the shortest path to remove scales.
\item\textbf{Robot direction and speed}: At time step $t$, find the union of all safe zones of all objects, $SZ_{t+1}=\bigcup_{i=1}^NSZ^i_{t+1}$. Then find the area $Env\setminus SZ_{t+1}$. For the direction of the robot, choose the point $Q\in Env\setminus SZ_{t+1}$ such that the length of the shortest path from $x$ to $B$ has the smallest value 
$$Q=\argmin_{x\in Env\setminus SZ_{t+1}}\left\{|Shortest\_Path (x,B)|~\mid~Straight\_line (act_t,x)\cap SZ_{t+1}=\emptyset\right\},$$
where $Straight\_line (act_t,Q)$ is the straight line connecting the points $act_t$ and $Q$. Note that the velocity value of the robot is constrained to $\mathrm{v}$ and the point $Q$ may not be reached in one time step. Thus, the robot either moves toward point $Q$ at the maximum velocity $\mathrm{v}$ for one time step or it can travel to $Q$ at the velocity 
$$v^r_{t+1}=\frac{|Shortest\_Path (act_t,Q)|}{\text{length of a time step}},$$
where $|Shortest\_Path (act_t,Q)|$ is a straight line.

So the state of the robot will be 
$$(orien_{t+1}^i, dist_{t+1}^r,d_{v^r_{t+1}},1,v^r_{t+1},1),$$ 
where $d_{v^r_{t+1}}$ is the angle of the line $Straight\_line(act_t,Q)$ with the $x$-axis in gradient scale and 
$$v^r_{t+1}=\min\left\{\mathrm{v},\frac{|Shortest\_Path (act_t,Q)|}{\text{length of a time step}}\right\}.$$
Note that after each time step, the current position of the robot and the shortest path should be updated. Since, by construction, for every two consecutive time steps $t$ and $t+1$ we have
$$0\leq|Shortest\_Path (act_{t+1},B)|\leq|Shortest\_Path (act_{t},B)|$$
the method converges. 

The way the robot's direction works at each time step allows it to find the optimal path, which may use a different route than the initial shortest path from the initial position to the robot's target position.
\item\textbf{Next position}: If the state of the object $i$ at time step $t$ is given by \eqref{eq:eq1}, then the next position $(orien_{t+1}^i, dist_{t+1}^i)$ can be determined as follows:
$$\min\{orien_t^i,d_{v^i_t}\}\leq orien_{t+1}^i=\arctan(\frac{Y}{X})\leq\max\{orien_t^i,d_{v^i_t}\}$$
and 
$$dist_{t+1}^i=\sqrt{X^2+Y^2},$$
where
$$X=dist_{t}^i\cos(orien_t^i)+v^i_t\cos(d_{v^i_t})$$
and 
$$Y=dist_{t}^i\sin (orien_t^i)+v^i_t\sin (d_{v^i_t}).$$
\item\textbf{Velocity and direction pattern}: Suppose that at each time step $t$, the past $H$ time steps of the states of all $N$ objects are known. This means 
$$(orien_s^i, dist_s^i,d_{v^i_s},P(d_{v^i_s},\-),v^i_s,P(v^i_s,\-)),~~s=t-H,\ldots,t-1.$$
We want to make a prediction of the directions $d_{v^i_s}$ and the velocity values $v^i_s$ for $s\geq t$. The following values can be natural choices:
\begin{itemize}
\item\textbf{Moving average}: For the future time step $t+k$, the velocity value of the object $i$ at time $t+k$ will be 
$$v^i_{t+k}=mean(\{v^i_{s}\mid s=t+k-H,\ldots,t+k-1\})$$
and 
$$d(v^i_{t+k})=mean(\{d(v^i_{s})\mid s=t+k-H,\ldots,t+k-1\}).$$
The advantage of this method is that the velocity values and directions change over time. However, it also has the disadvantage of becoming a linear function without updating if the number of time steps is long enough. This is a simple auto-regressive method for forecasting, see \cite{montgomery:2011}.
\item\textbf{Naive auto-regressive}: 
For the future time step $t+k$, the velocity value of the object $i$ at time step $t+k$ is determined as follows: Let 
$$W=(1,\frac{1}{2},\ldots,\frac{1}{2^{H-1}}).$$
Define
$$w=(w_{t+k-H},\ldots,w_{t+k-1}),$$
where $w_{t+k-u}=\frac{W_u}{\sum_{j=1}^{H}W_j}$ are $H$ weights for the past states.
Then
$$v^i_{t+k}=\sum_{j=1}^Hw_{t+k-j}v^i_{t+k-j}$$
and 
$$d(v^i_{t+k})=\sum_{j=1}^Hw_{t+k-j}d(v^i_{t+k-j}).$$
Velocity value and direction are related to past velocity values and directions. However, past velocity values and directions must be weighted, i.e., the next velocity value and direction are more related to the first immediate past velocity value and direction than the velocity values and directions in the time steps before it. 

The advantage of this method over the previous one is that the velocity values and directions change in a more realistic way over time. However, it also has the disadvantage that, without updating, the motion strategy is deterministic after a sufficiently long number of time steps, and also, the weights for the directions and velocity values are always the same all the time.
\item\textbf{Generalized auto-regressive}: For future time step $t+k$, the velocity value of object $i$ at time $t+k$ is determined as follows: Let 
$$w^u=(w^k_{t+k-H},\ldots,w^k_{t+k-1}),$$
for $u=1,2$, be $H$ random variables in the interval $(0,1)$ such that for $u=1,2$, $\sum_{j=1}^Hw^k_{t+h-j}=1$ and 
$$w^u_{t+k-H}\leq\ldots\leq w^u_{t+k-1}.$$
 $w^u_{t+k-j}$'s are weights. Let
$$v^i_{t+k}=\sum_{j=1}^Hw^1_{t+k-j}a^i_{t+k-j}$$ 
and 
$$d(v^i_{t+k})=\sum_{j=1}^Hw^2_{t+k-j}d(v^i_{t+k-j}).$$
The advantage of this method over the previous one is that the velocity values and directions change more realistically over time since the lists of probabilities are determined randomly. However, it also has the disadvantage that without additional information, some states may be invalid as future states of object $i$. This is a more general auto-regressive method for forecasting, see \cite{montgomery:2011}. 

We could also introduce random noise at each step by adding weighted multipliers of white noise, see \cite{montgomery:2011}.
\end{itemize}
In all the above natural possibilities, the position of the object $i$ at time $t+k$ can be determined using the step (Next position).
\item\textbf{State update}: When at time step $t$ the states of all objects are observed, we need to update the states 
$$(orien_t^i, dist_t^i,d_{v^i_t},1,v^i_t,1),~~ \forall i=1,\ldots,N.$$
The values $orien_t^i$ and $dist_s^i$ are updated by the current coordinate of the object $i$ at time $t$, and the direction and velocity value $d_{v^i_s}$ and $v^i_s$ will be updated by the current direction and velocity of the object $i$. Since the current values are observed, the probability densities of errors in the direction and velocity value of all objects can be avoided. Based on these new observations, all predictions for the states in the future should be updated using the methods described above by replacing the predicted states at time step $t$ with the actual values of the states at time step $t$. Note that if the current time step is $t$, this means that at time step $s$ with $s<t$, all states are already updated with the actual observed states.
\item\textbf{High risk scenarios (Run away)}: Let at time step $t$ $C_r(orien_t^r, dist_t^r)$ be the set of all points on the map that the robot can move from its current position, red circle in Figures \ref{f3p} and \ref{f4p}. Suppose that multiple objects, $i_1,\ldots,i_n$, are moving towards the robot such that 
$$(orien_t^r, dist_t^r)\in SZ^{i_j}_{t+1},~~\forall j=1,\ldots,n.$$
This means that the sets
$$\left\{|Shortest\_Path (x,B)|~\mid~Straight\_line (act_t,x)\cap SZ_{t+1}=\emptyset\right\},~~\forall x\in Env\setminus SZ_{t+1}$$
are empty sets for all $x$. In such cases, the next state of the robot can be determined as follows:
\begin{itemize}
\item\textbf{With skip zone}: Assume that 
$$D=C_r(orien_t^r, dist_t^r)\setminus\left(\bigcup_{j=1}^nSZ^{i_j}_{t+1}\right)\neq\emptyset.$$
The set $D$ is called the skip zone because if the robot moves to a point in $D$, it avoids possible collisions. For this case, find
$$Q=\argmin_{x\in D}\left\{|Shortest\_Path (x,B)|\right\}.$$ 
Note that the velocity value of the robot is constrained to $\mathrm{v}$ and the point $Q$ can be reached in a single time step. Thus, the robot moves towards the point $Q$ with the velocity value
$$v^r_{t+1}=\frac{|Shortest\_Path (act_t,Q)|}{\text{length of a time step}}\leq\mathrm{v}.$$

So the state of the robot will be 
$$(orien_{t+1}^i, dist_{t+1}^r,d_{v^r_{t+1}},1,v^r_{t+1},1),$$ 
where $d_{v^r_{t+1}}$ is the angle of the line $Straight\_line(act_t,Q)$ with the $x$-axis in gradient scale, $(orien_{t+1}^i, dist_{t+1}^r)$ is the polar coordinate of $Q$, and
$$v^r_{t+1}=\frac{|Shortest\_Path (act_t,Q)|}{\text{length of a time step}}.$$
Figure \ref{f3p}, illustrate the scenario with skip zone.
\begin{figure}[!h]\centering
\includegraphics[width=0.8\linewidth]{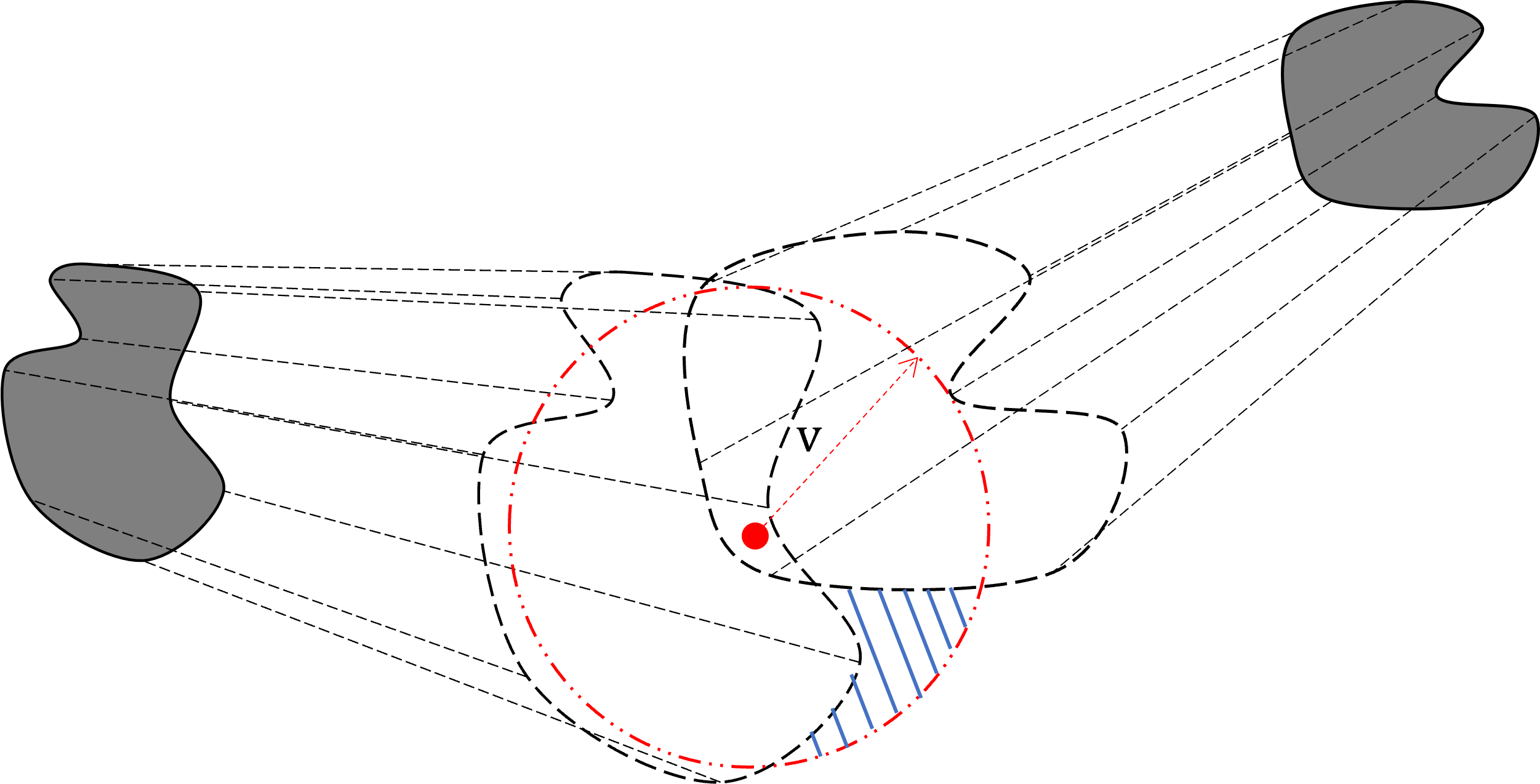}
\caption{Intuition to avoid collision, for the case with skip zone. The red bullet is the current position of the robot, the red vector indicates the maximum velocity value of the robot, and the red circle with the double dotted and dashed line indicates the set of positions that the robot can move with its limited velocity value. The blue dashed area is the skip zone $D$.}
\label{f3p}
\end{figure}
\item\textbf{Without skip zone}: Assume that 
$$C_r(orien_t^r, dist_t^r)\setminus\left(\bigcup_{j=1}^nSZ^{i_j}_{t+1}\right)=\emptyset.$$
See figure \ref{f4p} for such a case. 
\begin{figure}[!h]\centering
\includegraphics[width=0.8\linewidth]{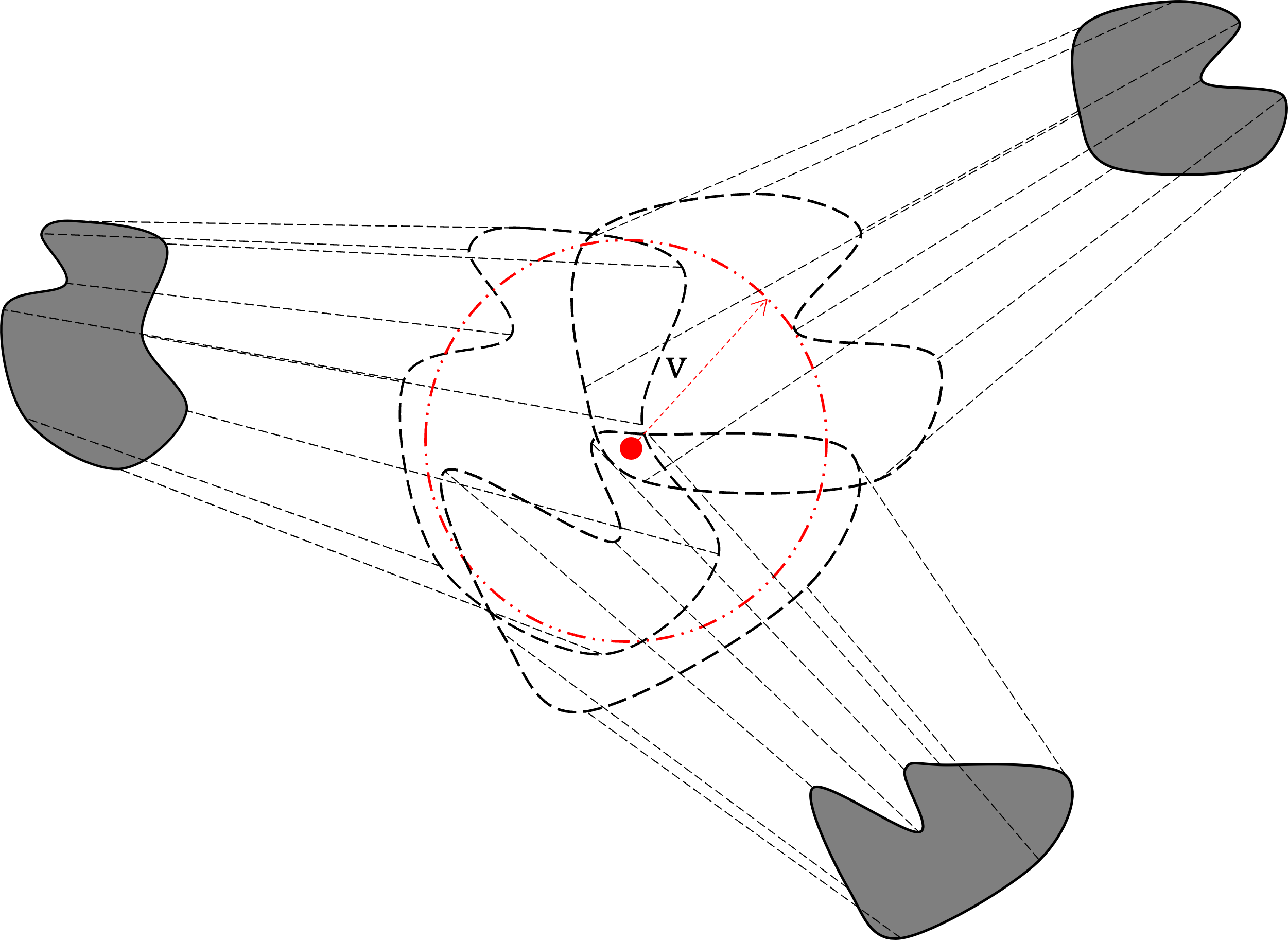}
\caption{The case without skip zone. The red bullet is the robot's current position, the red vector shows the robot's maximum velocity value, and the red circle with the double dotted and dashed line shows the set of positions the robot can travel to with its limited velocity value.}
\label{f4p}
\end{figure}
For this case, find the set
$$M=\argmin_{x\in C_r(orien_t^r, dist_t^r)}\left\{\sum_{j=1}^n\left(P(d_{v^{i_j}_t},x)+P(v^{i_j}_t,x)\right)\right\}.$$
The set $M$ is the set of all points that the robot can move to in a single time step with the least probability that objects $i_1,\ldots,i_n$ move to those points. Then, as before
$$Q=\argmin_{x\in M}\left\{|Shortest\_Path (x,B)|\right\}.$$ 
Note that the velocity value of the robot is constrained to $\mathrm{v}$ and the point $Q$ can be reached in a single time step. Thus, the robot moves towards the point $Q$ with the velocity value
$$v^r_{t+1}=\frac{|Shortest\_Path (act_t,Q)|}{\text{length of a time step}}\leq\mathrm{v}.$$

So the state of the robot will be 
$$(orien_{t+1}^i, dist_{t+1}^r,d_{v^r_{t+1}},1,v^r_{t+1},1),$$ 
where $d_{v^r_{t+1}}$ is the angle of the line $Straight\_line(act_t,Q)$ with the $x$-axis in gradient scale, $(orien_{t+1}^i, dist_{t+1}^r)$ is the polar coordinate of $Q$, and
$$v^r_{t+1}=\frac{|Shortest\_Path (act_t,Q)|}{\text{length of a time step}}.$$
\end{itemize}

Note that throughout the formulation the robot is considered as a single point (centroid of the robot, $act_t$) at time step $t$. To generalize the formulation, assume that $OCP_t^r\subset Env$ are the positions in the environment that the robot occupies at time step $t$. Then all formulations involving the motion of the robot from $act_t$ to a point $w\in Env$ should be replaced by the mapping 
$$\delta:OCP_t^r\rightarrow B(w,act_t,OCP_t^r),$$
which maps $act_t$ to $w$ with direction $d(v)$ (direction from $act_t$ to $w$) and velocity $v$ (the distance between $act_t$ and $w$ divided by the length of the time step), and $y\in OCP_t^r$ to the point $y_1\in B(w,act_t,OCP_t^r)$, where the line connecting $y$ and $y_1$ makes an angle $d(v)$ with the $x$-axis and the distance between $y$ and $y_1$ is equal to $v$. Then the whole formulation should be applied to the $(u,v)\ in OCP_t^r\times\delta(OCP_t^r)$, instead of only to $(act_t,w)$.
\end{itemize}

\section{Conclusion}
We have developed a simple model that allows optimal path planning in an unknown dynamic environment, minimizing the total distance the robot travels and the total time to complete a given task, while avoiding collisions. To avoid collisions, we consider no bounding box around obstacles to further minimize the distance traveled by the robot. For moving obstacles, we perform a simple prediction of the future motion of the obstacles and the motion of the robot changes accordingly. Over time, the robot updates the state of the moving objects to better estimate their future motion. 

For the future work, to better estimate the next motion of the obstacles, we will replace the simple estimation with a better prediction method, such as more complex time series prediction and forecasting methods.

\section*{Acknowledgment}

This work was supported by operation Centro-01-0145-FEDER-000019-C4-Centro de Compet\^{e}ncias em Cloud Computing, cofinanced by the European Regional Development Fund (ERDF) through the Programa Operacional Regional do Centro (Centro 2020), in the scope of the Sistema de Apoio \`{a} Investiga\c{c}\~{a}o Cientif\'{i}ca e Tecnol\'{o}gica - Programas Integrados de IC\&DT. This work was supported by NOVA LINCS (UIDB/04516/2020) with the financial support of FCT-Funda\c{c}\~{a}o para a Ci\^{e}ncia e a Tecnologia, through national funds.

\bibliographystyle{unsrt}
\bibliography{sample}

\end{document}